\documentclass{article}

     \PassOptionsToPackage{numbers, compress}{natbib}

\usepackage[preprint]{neurips_2021}

\usepackage[utf8]{inputenc} 
\usepackage[T1]{fontenc}    
\usepackage{hyperref}       
\usepackage{url}            
\usepackage{booktabs}       
\usepackage{amsfonts}       
\usepackage{nicefrac}       
\usepackage{microtype}      
\usepackage{xcolor}         
\usepackage{subfigure}
\usepackage{graphicx}
\usepackage{comment}
\usepackage{wrapfig}
\usepackage{multirow}
\usepackage{setspace}
\usepackage{colortbl}
\usepackage{amsmath,amssymb} 
\usepackage{algorithmic}
\usepackage[ruled,vlined]{algorithm2e}  

\usepackage{url}

\title{Gradient Concealment: Free Lunch for Defending Adversarial Attacks}

\author{%
  Sen Pei\\
  NLPR, Institute of Automation\\
  Chinese Academy of Science\\
  \texttt{peisen2020@ia.ac.cn} \\
   \And
  Jiaxi Sun\\
  NLPR, Institute of Automation\\
  Chinese Academy of Science\\
  \texttt{sunjiaxi2020@ia.ac.cn} \\
  \AND
  Xiaopeng Zhang \\
  NLPR, Institute of Automation \\
  Chinese Academy of Science \\
  \texttt{xpzhang@nlpr.ia.ac.cn} \\
  \And
  Gaofeng Meng \\
  NLPR, Institute of Automation \\
  Chinese Academy of Science \\
  \texttt{gfmeng@nlpr.ia.ac.cn} \\
}

\begin{document}

\maketitle

\begin{abstract}
Recent studies show that the deep neural networks (DNNs) have achieved great success in various tasks. However, even the \emph{state-of-the-art} deep learning based classifiers are extremely vulnerable to adversarial examples, resulting in sharp decay of discrimination accuracy in the presence of enormous unknown attacks. Given the fact that neural networks are widely used in the open world scenario which can be safety-critical situations, mitigating the adversarial effects of deep learning methods has become an urgent need. Generally, conventional DNNs can be attacked with a dramatically high success rate since their gradient is exposed thoroughly in the white-box scenario, making it effortless to ruin a well trained classifier with only imperceptible perturbations in the raw data space. For tackling this problem, we propose a plug-and-play layer that is training-free, termed as \textbf{G}radient \textbf{C}oncealment \textbf{M}odule (GCM), concealing the vulnerable direction of gradient while guaranteeing the classification accuracy during the inference time. GCM reports superior defense results on the ImageNet classification benchmark, improving up to 63.41\% top-1 attack robustness (AR) when faced with adversarial inputs compared to the vanilla DNNs. Moreover,  we use GCM in the CVPR 2022 Robust Classification Challenge \footnote{https://aisafety.sensetime.com}, currently achieving \textbf{2nd} place in Phase II with only a tiny version of ConvNext. The code will be made available.
\end{abstract}

\section{Introduction}
\label{intro}
\begin{figure}[htpb]
\begin{center}
\centerline{\includegraphics[width=0.9\columnwidth]{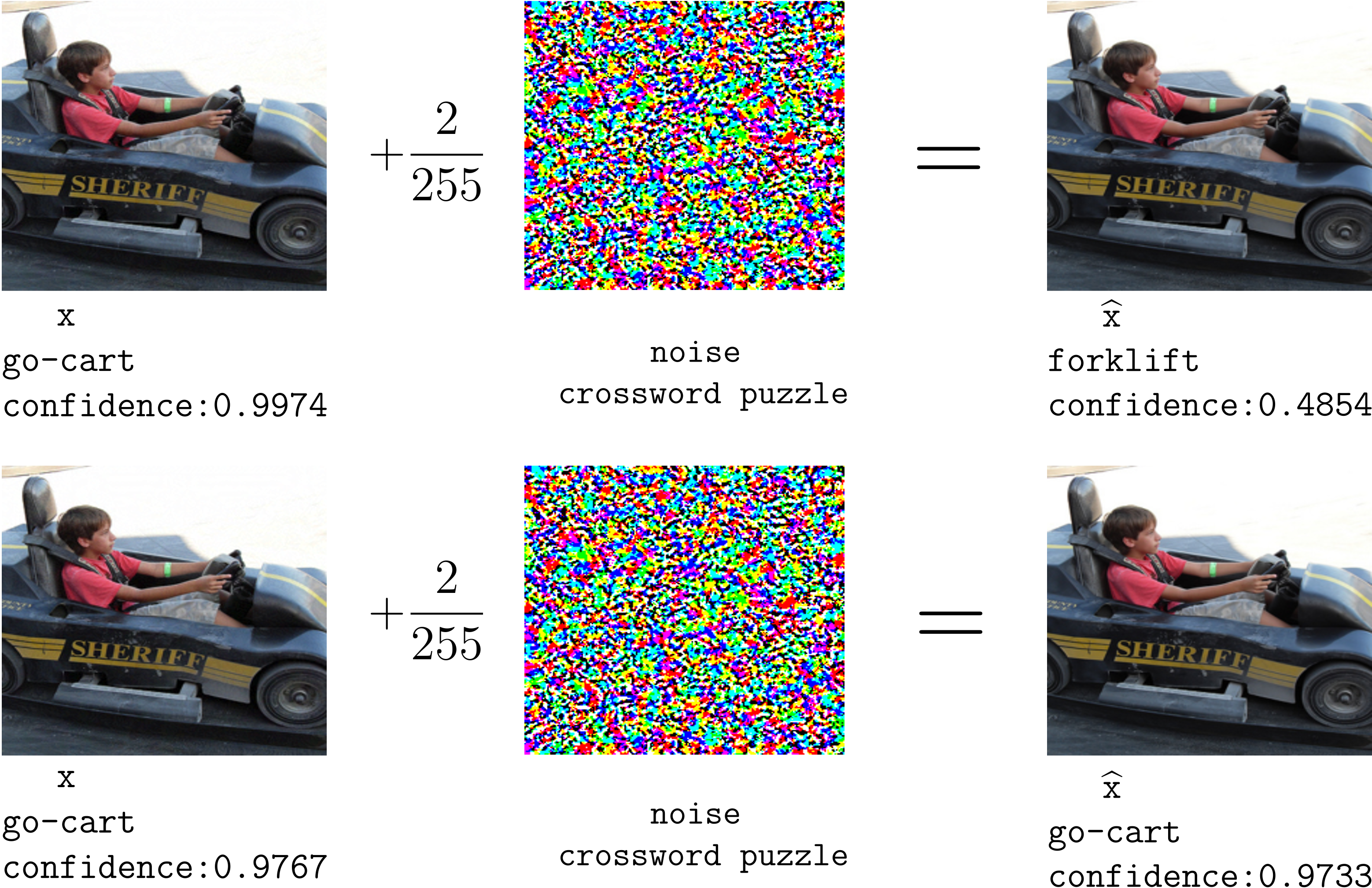}}
\caption{\textbf{Attacking DNNs with FGSM \cite{1}.} Top: Vanilla ResNet50 \cite{7} trained on ImageNet 2012 dataset \cite{26}. Bottom: Identical ResNet50 with our proposed GCM. The left images are correctly classified as go-cart. The center images are adversarial perturbations generated by FGSM. We randomly choose \texttt{crossword puzzle} as the target class after being attacked, and the perturbations are obtained by minimizing the cross entropy loss between the output of ResNet and the target one-hot label. It can be seen that the perturbed image in the first line is misclassified as forklift with a confidence up to 0.4854 by the vanilla ResNet50. For contrast, with our proposed GCM, the identical ResNet50 can defend these attacks effortlessly.}
\label{fig:illustration}
\end{center}
\vskip -0.3in
\end{figure}

The studies towards model robustness are highly valued in security-critical applications where the reliability of DNNs against unknown inputs must be treated carefully\cite{4}. From the current research in \cite{5,6,7,8}, DNNs have been proved to generalize well when the input data are clean and drawn \emph{i.i.d.} from the same distribution as the training set. However, in the open world scenario, DNNs can be attacked maliciously with visually unnoticeable perturbations, resulting in severe performance decay and security risk (see Figure~\ref{fig:illustration}). 
For remedying DNNs from vulnerability when being attacked, previous works in \cite{1,2,3,9,10,11,12} have made a great contribution to this line of research, trying to promote DNNs much more robust in the presence of perturbed images. Unfortunately, these methods only focus on some specific attacks which can not be generally used in various applications, and as a result, leave potential risks in complicated unknown situations. In this paper, we propose to conceal the vulnerable direction of the classifier's gradient for improving its reliability, protecting DNNs from gradient based attacks.

Currently, there is now a sizable body of work proposing various attack schemes to ruin the superior performance of DNNs, and unfortunately, most adversarial attacks meet their need. Generally speaking, the adversarial attack can be viewed in two manners, which are the black-box attack and the white-box attack. In the case of the former fashion, noisy images generated with only a limited (or without) knowledge of the model are fed into the classifier for testing, and the data augmentation methods can help to deal with this situation \cite{13}. By comparison, the white-box attack is much more aggressive since it has access to the complete details of the target model, including its parameter values, architecture, training method, and in some cases its training data as well. In all but a few cases, white-box attacks are gradient based attack methods which can be roughly divided into two categories: the single-step attack and the iterative attack. As in \cite{3}, the perturbed images generated by iterative methods such as DeepFool~\cite{14} may easily get over-fitted to the target neural network, resulting in stronger but less transferable attacks. On the contrary, the single-step attack such as FGSM~\cite{1} narrowly suffers from the aforementioned issue, but its generated perturbations may not be aggressive enough to fool the target classifier. All told, concealing the gradients of DNNs is one of the key solutions to prevent classifiers from being attacked, leading to both stable performance and better generalization ability in the presence of unknown adversarial attacks. 

Towards this end, we propose using GCM to hide the direction of gradients for mitigating adversarial effects, protecting DNNs from malicious attacks. GCM is a plug-and-play module placed in an arbitrary layer of DNNs which generates marginal magnitude and enormous gradient, resulting in a negligible change in feature maps while significant disturbance in the sign of gradient. The proposed GCM is parameter-free, introducing no classification performance decay during both the training and inference time. The main contributions of this work are summarized as follows:
\vspace{-0.5em}
\begin{itemize}
\item A parameter-free module termed GCM is proposed for defending adversarial attacks without costly training, increasing up to 63.41\% attack robustness (AR) in the presence of gradient based attacks compared to previous methods.

\item We test the proposed GCM on ImageNet classification benchmarks with ResNet~\cite{2}, WideResNet~\cite{27}, DenseNet~\cite{8}, EfficientNet~\cite{28} and different transformer architectures such as ViT~\cite{29} and Swin-Transformer~\cite{30}. These sufficient results suggest that GCM improves the performance of adversarial defense stably, making it more reliable to apply neural networks in security-critical areas.

\item Unlike prior works, we pose a brand new perspective to explore the solution of adversarial defense in deep learning, leading to defense schemes with better generalization ability.
\end{itemize}

\section{Related Work}
In this section, we give an overview of adversarial attacks and some current defense methods for improving the robustness of deep neural networks.

\noindent
\textbf{Adversarial Attacks} are severe threats for deploying machine learning systems in the security-critical scenario. In \cite{15}, adversarial examples were first noticed. It is found that adding visually undetectable noise to the clean images can result in the misclassification of deep neural networks. With $L_2$ distance metric, these adversarial images can be obtained through box-constrained L-BFGS \cite{24}. Later, in \cite{1}, a single-step attack approach named Fast Gradient Sign Method (FGSM) is proposed to produce adversarial examples efficiently using the sign of gradient under the constraint of $L_{\infty}$ distance metric. In \cite{16}, FGSM is refined in an iterative manner taking multiple aggressive steps for attacking, reporting superior results compared to the vanilla FGSM. PGD \cite{20} is also an iterative attack technique using the first-order gradient of the target classifiers. It clips the perturbations in each iteration within a $\epsilon$ ball. Jacobian-based Saliency Map Attack (JSMA) introduced in \cite{17} is a greedy attack method optimized with $L_0$ distance. The proposed saliency map models the impact each pixel has on the corresponding classification result, and JSMA modifies the pixel with the highest saliency for improving the likelihood of the desired label. Similarly, \cite{19} finds that modifying only one pixel can also destroy the vulnerable deep neural networks, raising more concern about AI Safety. In \cite{14}, DeepFool proposes generating closer adversarial examples under the constraint of $L_2$ distance metric, providing a better understanding of adversarial effects. C\&W \cite{18} aims to find some imperceptible noise that minimizes the difference between the perturbed image and its clean counterpart under some desired distance metric while fooling the well-trained classifiers. C\&W is an iterative gradient based attack method with an extremely high success rate for ruining current defense approaches such as FGSM regularization \cite{1} and defensive distillation \cite{9}. Alongside, studies in \cite{21,22,23} focus on the transferability of adversarial examples. UAP \cite{21} is a systematic algorithm for computing architecture-agnostic perturbations, revealing the geometric correlation between different parts of the decision boundary. More recently, a plug-and-play module named MGAA \cite{23} proposes generating universal noise iteratively with the meta-train step and the meta-test step, improving the cross-model transferability of these perturbations. Different from the aforementioned approaches, AdvDrop \cite{25} crafts adversarial samples by dropping existing details of images, making defense a much more difficult task.

\noindent
\textbf{Adversarial Defense} is an opposite research direction to mitigate the adversarial effects. The progress in defending against attacks is not such significant compared to the generation of adversarial examples. In \cite{31}, feature squeezing is proposed to detect adversarial examples by comparing DNN model's prediction on the original input with that of squeezed inputs. Defense-GAN \cite{32} proposes using a generator to denoise the adversarial examples by minimizing the reconstruction error between these perturbed images and the synthetic images generated by the aforementioned generator. Unlike previous adversarial training schemes such as TRADES \cite{40}, \cite{33} generates adversarial images for training through feature scattering in latent space, avoiding label leaking. \cite{34} claims that the vulnerability of DNNs is caused directly by the non-robust features in images, and capturing these features can reduce the success rate of adversarial attacks. Defensive distillation proposed in \cite{35} finds that model distillation can also reduce the effectiveness of adversarial samples on DNNs. \cite{36} proposes a recursive generator that can produce much stronger adversarial perturbations for training, revealing the vulnerability of the target classifier. \cite{37} argues that the close proximity of different class samples in the learned feature space causes the poor performance of DNNs in the presence of imperceptible adversarial noise, and thus restricting the hidden space of DNNs help to defend against these attacks. Sparse coding \cite{39} finds that projecting images into its proposed quasi-natural image space can remove the adversarial perturbations, and this method is attack-agnostic which has good generalization ability. HGD \cite{39} is a denoiser that uses a loss function defined as the difference between the target classifier's outputs activated by the clean image and its corresponding denoised image, removing adversarial perturbations adaptively. Different from the aforementioned training schemes, the randomization layer proposed in \cite{41} is a post-processing method without trainable parameters. It invalidates the adversarial attacks by changing the attacked location and pixel values through image resizing and random padding. Compared to the above techniques, the proposed training-free GCM has better generalization ability and much stronger defense capability, making gradient based attacks ineffective.

\section{Approach}
\label{approach}
At the beginning of this section, we detail the adversarial attack methods used in the following experiments. Immediately, we elaborate the proposed \textbf{G}radient \textbf{C}oncealment \textbf{M}odule formally.

\subsection{Gradient based Adversarial Attacks}
Before introducing the proposed GCM, we give an overview of gradient based adversarial attack methods formally. Suppose $x$ and $y$ are the legitimate (\emph{i.e.}, clean) image without perturbations and its corresponding ground truth label. Let $L(x,y,\theta)$ denote the loss function of target classifier parameterized by $\theta$. The perturbed image $\hat{x}=x+r$ is obtained through different attack methods where $r$ indicates the adversarial noise with the constraint of $||r||_p\le \epsilon$. Here, we use $||\cdot||_p$ to represent the $L_p$ norm. The desired label after being attacked is $\hat{y}$. We consider three adversarial attacks in the white-box fashion which are FGSM~\cite{1}, PGD~\cite{20} and C\&W~\cite{18} respectively.

\noindent
\textbf{Fast Gradient Sign Method (FGSM) \cite{1}} is a basic single-step attack method using the sign of gradient. Usually, FGSM is treated in two fashions which are target-attack and untarget-attack. The former case desires to change the predicted label of input $x$ to some specific target labels. The latter case only wants to make the model misclassify the perturbed images. Formally, the untarget-attack is built as follows with the constraint of infinite norm $\epsilon$:
\begin{equation}
\hat{x}=x+\epsilon \cdot \mathrm{sgn}(\nabla_x L(x,y,\theta))
\label{eq:fgsm-untarget}
\end{equation}
Opposite to the above equation, target-attack can be established using gradient descend method as:
\begin{equation}
\hat{x}=x-\epsilon \cdot \mathrm{sgn}(\nabla_x L(x,\hat{y},\theta))
\label{eq:fgsm-target}
\end{equation}

\noindent
\textbf{Projected Gradient Descent Method (PGD) \cite{20}} proposes an unified framework for generating adversarial examples together with the adversarial training process. The objective of PGD is to solve a min-max optimization problem shown as follows:
\begin{equation}
\min_\theta \rho(\theta)=\mathbb{E}_{(x,y)\sim \mathcal{D}} \;\max_{||r||_p \leq \epsilon}L(x+r,y,\theta)
\label{eq:pgd}
\end{equation}
where $\mathcal{D}$ is the training data, $\theta$ is the parameters of the target classifier and $p$ is the order of norm.

\noindent
\textbf{C\&W \cite{18}} is an iterative attack method. With an auxiliary variable $w$, it finds adversarial perturbations $r$ as:
\begin{equation}
r=\frac{1}{2}(\tanh(w)+1)-x
\label{eq:cw-noise}
\end{equation}
where the perturbed image $\hat{x}=x+r=\frac{1}{2}(\tanh(w)+1)$. C\&W aims to minimize the $L_p$ norm of noise $r$ while making the model misclassify the aforementioned image $\hat{x}$. Formally, the unified loss function is in the following form~\cite{41}:
\begin{equation}
\min_{w} ||r||_p+c\cdot \max\{f(\hat{x})_y-f(\hat{x})_{\neq y},-k\}
\label{eq:cw-optimize}
\end{equation}
where $f(x)_y$ denotes the predicted confidence of $x$ matching with label $y$, and $k$ is a desired confidence gap between the ground truth label $y$ and any other misclassified labels.

\subsection{Gradient Concealment Module}
\begin{figure}[t]
\begin{center}
\centerline{\includegraphics[width=0.95\columnwidth]{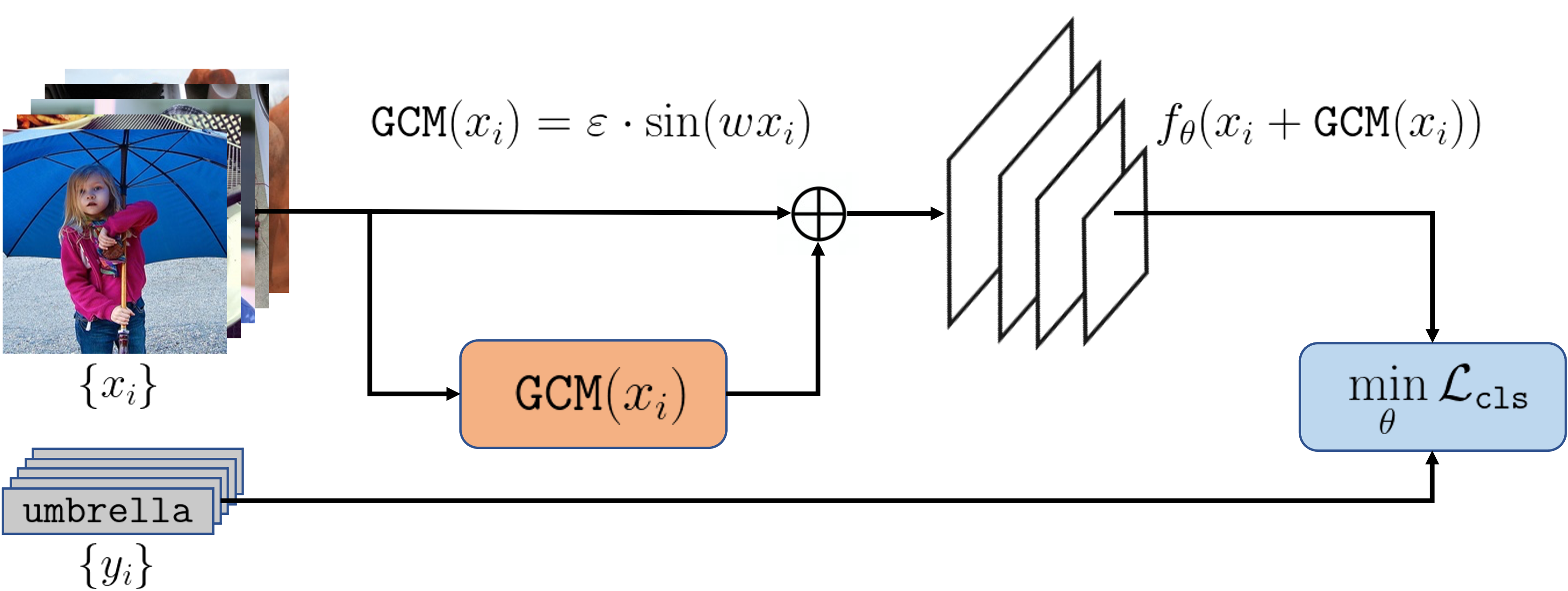}}
\caption{\textbf{The proposed GCM framework.} Gradient concealment module is a plug-and-play layer, and it can greatly reduce the magnitude of input data while generating enormous gradient. $w$ is the frequency of trigonometric function, and $\varepsilon$ is a extremely tiny weight. When setting $w \cdot \varepsilon$ far greater than 1, the vulnerable direction of classifier's gradient is concealed, providing stable and superior performance in the presence of adversarial attacks. $\{x_i\}$ and $\{y_i\}$ are the clean training images and labels, $f_{\theta}(\cdot)$ is a deep neural network parameterized by $\theta$ and $\mathcal{L}_{\texttt{cls}}$ is the category cross entropy loss.}
\label{fig:gcm}
\end{center}
\vskip -0.3in
\end{figure}

GCM is used to conceal the vulnerable gradient direction of the vanilla classifiers, making gradient based attacks completely ineffective. During the training phase, the classifier is trained as usual with the backpropagation method. In the inference time, one just needs to place the GCM at the top (or any other layers) of the trained neural networks. Since the magnitude of output from GCM is extremely minute so that no obvious change is introduced to the classification accuracy. 

\begin{wrapfigure}{r}{0.5\textwidth}
  \vspace{-15pt}
  \begin{center}
   \includegraphics[width=0.48\textwidth]{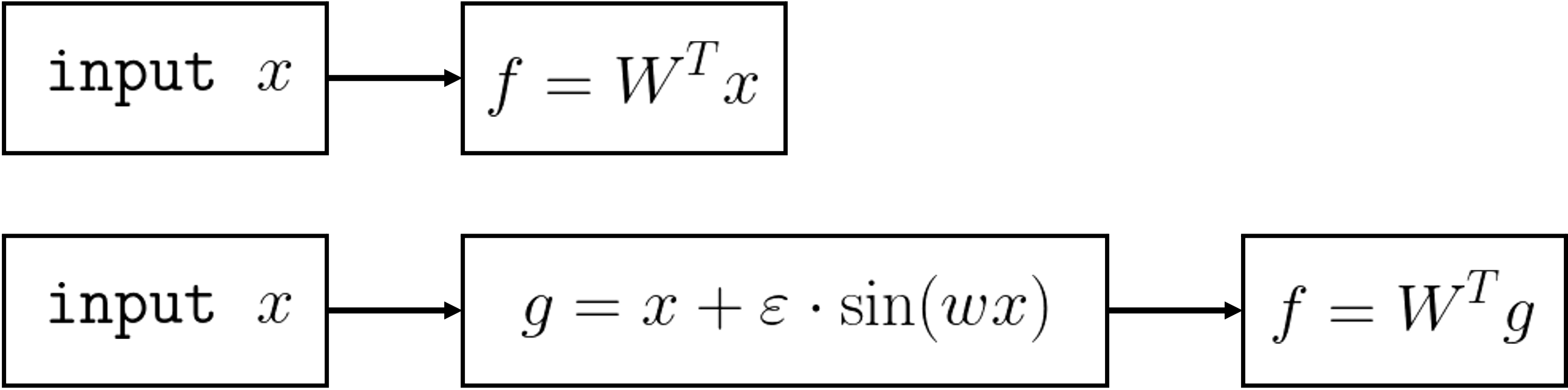}
  \end{center}
  \caption{\textbf{Calculation flow of the vanilla model and its GCM counterpart.} $f(\cdot)$ is the classifier and $g(\cdot)$ indicates the proposed GCM.}
  \label{fig:flow}
\end{wrapfigure}

Formally, let $x$ be the input image, and let $f(\cdot)$ denote the deep neural network. In Figure~\ref{fig:flow}, the upper part shows the calculation flow of conventional DNNs, and the lower half of the figure describes that of our proposed GCM. Since the parameter $\varepsilon$ is extremely small, the forward process is almost identical in both the vanilla model and its GCM counterpart. We consider the backward process in the attack scenario. For the vanilla model, we have:
\begin{equation}
(\frac{\partial \mathcal{L}}{\partial x})_{\texttt{vanilla}}=\frac{\partial \mathcal{L}}{\partial f}\cdot \frac{\partial f}{\partial x}
\label{eq:vanilla_gradient}
\end{equation}
where $\mathcal{L}$ denotes the category cross entropy loss. Similarly, the gradient that $\mathcal{L}$ towards input $x$ in GCM is obtained with chain rule:
\begin{equation}
(\frac{\partial \mathcal{L}}{\partial x})_{\texttt{GCM}}=\frac{\partial \mathcal{L}}{\partial f}\cdot \frac{\partial f}{\partial g}\cdot \frac{\partial g}{\partial x}\approx (\frac{\partial \mathcal{L}}{\partial x})_{\texttt{vanilla}}\cdot \frac{\partial g}{\partial x}=(\frac{\partial \mathcal{L}}{\partial x})_{\texttt{vanilla}}\cdot (1+\varepsilon w\cos(wx))
\label{eq:gcm_gradient}
\end{equation}
When setting $\varepsilon w$ far greater than 1, the last term in Eq.(\ref{eq:gcm_gradient}) is dominated by $\varepsilon w\cos(wx)$, and therefore, concealing the original gradient's direction shown in Eq.(\ref{eq:vanilla_gradient}). GCM has no trainable parameters which can be used during the inference time directly. At a high level, the overall pipeline of the proposed GCM is shown in Algorithm \ref{algo:gcm}.

\begin{algorithm}[htpb]
\label{algo:gcm}
    \caption{Gradient concealment module for defending adversarial attacks.}
    \LinesNumbered
    \KwIn{training data $\mathcal{D}$, classifier $f_{\theta}(\cdot)$ parameterized by $\theta$, $w$ and $\varepsilon$ for GCM}
    \KwOut{trained neural network $f_{\theta}(\cdot)$}
    \While {\texttt{Training}}{
			{Sample a batch data $\{x_i,y_i\}_{i=1:K}$ from $\mathcal{D}$\;
			Get the corresponding output logits : $y_i^{*}=f_{\theta}(x_i)$\;
			Calculate the cross entropy loss $\mathcal{L}_{\texttt{cls}}$ between $(y_i^{*},y_i)_{i=1:K}$\;
			Update $\theta$ with gradient descent method.}
}

    \While {\texttt{Inference}}{
            Sample $\{x,y\}$ from the testing set\;
			Cascade : $g_{\theta}(x) = [x+\texttt{GCM}(x), f_{\theta}(\cdot)]$\;
			Adversarial attack : $\hat{x}=\texttt{attack}(x,g_{\theta}, y)$\;
			Evaluation : $\hat{y}=g_{\theta}(\hat{x})$.
}
\end{algorithm}

\section{Experiments}
\label{exp}
Starting with the experimental setup, we tell the evaluation metrics and report the adversarial defense results on ImagNet~\cite{26} and CIFAR-10~\cite{46} with different model architectures including CNN-based methods and attention-based methods immediately. Following this, we detail the ablation study and give some auxiliary visualization results.

\subsection{Experimental Setup}
\noindent
\textbf{Dataset.} We use the whole testing set from ImageNet~\cite{26} 2012 for evaluation. This testing set contains 50k images in 1k categories. The short edge of all images is resized to 256 first, maintaining the aspect ratio, and then the testing inputs are obtained through cropping at the center of images. The size of inputs after pre-processing is $224\times 224\times 3$ with RGB channels. Besides, we also report the defense results on CIFAR-10~\cite{46} compared to different defending methods, and no modifications are introduced to these images. The size of inputs is $32\times 32\times 3$ with RGB channels.

\noindent
\textbf{Model Architectures.} We employ sufficient model architectures to test the reliability and generalization performance of the proposed GCM. For convolution based models, we adopt ResNet-50~\cite{7}, WideResNet-50~\cite{27}, DenseNet-121~\cite{8}, ShuffleNetV2~\cite{42}, EfficientNet-B4~\cite{28} and MobileNetV3~\cite{43} from 2016 to 2019. For the attention based architecture which is currently a hot research topic, we use DeiT-S~\cite{44}, ViT-B/16~\cite{29} and Swin-Transformer-S~\cite{30} for comparison. All these models are trained on ImageNet 2012 training set, and their weights can be downloaded at this web page \footnote{http://robust.art/source}. The details of training process are stated clearly in RobustART~\cite{45}.

\noindent
\textbf{Evaluation Metrics.} We use the top-1 classification accuracy (ACC) and the attack robustness (AR)~\cite{45} to evaluate the defense ability of target classifiers. To formalize these metrics, let $\mathcal{D}=\{(x_1,y_1),(x_2,y_2),...,(x_N,y_N)\}$ denote the testing set, and let $f_{\theta}(\cdot)$ indicate the classifier parameterized by $\theta$. The corresponding perturbed image of $x_i$ is $\hat{x_i}$. With these notations, the classification accuracy is defined as:
\begin{equation}
\texttt{ACC}=\frac{1}{N}\sum_{i=1}^{N}\mathbb{I}(f_{\theta}(x_i)=y_i)
\label{eq:acc}
\end{equation}
where $\mathbb{I}(\cdot)$ is a binary indicator function whose value equals to 1 if and only if the inner condition is true. It is meaningless to attack the images which are already misclassified by the target model. Consequently, attack robustness (AR) only concerns the defense performance of model on the correctly classified sub testing set. AR is formulated as follows:
\begin{equation}
\texttt{AR}=\frac{\sum_{i=1}^{N}\mathbb{I}(f_{\theta}(x_i)=y_i)\cdot \mathbb{I}(f_{\theta}(\hat{x_i})=y_i)}{\sum_{i=1}^{N}\mathbb{I}(f_{\theta}(x_i)=y_i)}
\label{eq:ar}
\end{equation}

\noindent
\textbf{Adversarial Attack Methods.} We consider three adversarial attack methods which are FGSM~\cite{1}, PGD~\cite{20} and C\&W~\cite{18}. For single-step adversarial attack method FGSM, the perturbations $r$ are constrained with $L_{\infty}$ norm. For PGD, we report results under the constraint of $L_1$ norm, $L_2$ norm and $L_{\infty}$ norm respectively. For iterative attack method C\&W, we set the binary search steps to 10 with a learning rate 1e-2. The iteration steps of optimization is limited to 10.

\subsection{Main Results of Adversarial Defense}
We report the adversarial defense results on CIFAR-10~\cite{46} and ImageNet~\cite{26} 2012 Testing Set. Top-1 classification accuracy (ACC) is used to evaluate the performance of classifiers on clean images, and attack robustness (AR) is used to measure the stability and reliability of models in presence of adversarial attacks.

\noindent
\textbf{Adversarial Defense on ImageNet~\cite{26}.} Table~\ref{tab:imagenet} shows the classification result and the attack robustness result on ImageNet. ACC on clean images indicates the top-1 classification accuracy of the following models on the ImageNet testing set. All other metrics are attack robustness (AR) evaluated only on the correctly classified testing images. For each model in the first half of Table~\ref{tab:imagenet}, we set a corresponding GCM counterpart for comparison. One can notice the superior performance of GCM clearly by comparing the groups with and without GCM. The hyper-parameters $w$ and $\varepsilon$ in GCM are set to 1e20 and 1e-8 respectively. In particular, with FGSM attack method under the constraint of $L_{\infty} \leq \frac{8}{255}$, GCM dramatically improves the AR metric up to 63.41\% compared to the vanilla ResNet. Specially, due to the relatively high time cost in C\&W attack, we randomly choose 20 images per category in the ImageNet testing set for evaluation.

\setlength{\tabcolsep}{2.2pt}
\begin{table}[htpb]\scriptsize
\begin{spacing}{1.35}
\begin{center}
\caption{\textbf{Adversarial defense results on ImageNet~\cite{26}.} It can be seen clearly that GCM significantly improves the defense ability towards gradient based attacks while maintaining the classification accuracy of the input images. The table in light gray background denotes the results obtained with GCM. Values below are attack robustness if not specified.}
\label{tab:imagenet}
\vskip -0.15in
\begin{tabular}{c|c|cc|cccccc|c}
\hline
\multirow{2}{*}{\begin{tabular}[c]{@{}c@{}}ImageNet~\cite{26}\\ 2012 Testing Set\end{tabular}} & \begin{tabular}[c]{@{}c@{}}ACC on\\ Clean Images\end{tabular} & \multicolumn{2}{c|}{FGSM~\cite{1}} & \multicolumn{6}{c|}{PGD~\cite{20}}                                   & \multicolumn{1}{c}{C\&W~\cite{18}}      \\
                                                                                     &                                                               & $L_{\infty} \frac{2}{255}$   & $L_{\infty} \frac{8}{255}$   & $L_1$ 400 & $L_1$ 1600 & $L_2$ 2.0 & $L_2$ 8.0 & $L_{\infty} \frac{2}{255}$ & $L_{\infty} \frac{8}{255}$ & \multicolumn{1}{c}{$L_2$ 8.0}\\\hline
ResNet-50 (2016)~\cite{7}                                                                      & 77.890                                                        & 33.560      & 31.773      & 3.286  & 0.012   & 2.128  & 0.010  & 1.450     & 0.000     & \multicolumn{1}{c}{0.211}          \\
WideResNet-50 (2016)~\cite{27}                                                                  & 78.212                                                        & 26.077      & 20.879      & 3.250  & 0.358   & 2.408  & 0.611  & 2.250     & 0.496     & \multicolumn{1}{c}{0.213}          \\
DenseNet-121 (2017)~\cite{8}                                                                   & 74.858                                                        & 20.016      & 16.815      & 1.130  & 0.035   & 0.587  & 0.053  & 0.408     & 0.056     & \multicolumn{1}{c}{0.116}          \\
ShuffleNetV2-x2.0 (2018)~\cite{42}                                                              & 72.286                                                        & 8.662       & 7.500       & 0.138  & 0.000   & 0.088  & 0.000  & 0.085     & 0.000     & \multicolumn{1}{c}{0.119}          \\
EfficientNet-B4 (2019)~\cite{28}                                                                & 71.520                                                        & 2.052       & 1.233       & 5.819  & 0.357   & 0.769  & 0.276  & 0.363     & 0.198     & \multicolumn{1}{c}{1.884}          \\
MobileNetV3-x1.4 (2019)~\cite{43}                                                               & 73.636                                                        & 10.568      & 9.245       & 0.399  & 0.000   & 0.105  & 0.000  & 0.152     & 0.000     & \multicolumn{1}{c}{0.236}          \\
DeiT-S (2021)~\cite{44}                                                                         & 79.900                                                       & 20.684      & 8.828       & 1.585  & 0.000   & 0.881  & 0.000  & 0.990     & 0.000     & 0.644                            \\
ViT-B/16 (2021)~\cite{29}                                                                       & 79.462                                                        & 31.640      & 15.862      & 3.522  & 0.003   & 1.929  & 0.000  & 2.511     & 0.000     & 0.902                              \\
Swin-Transformer-S (2021)~\cite{30}                                                             & 82.930                                                        & 25.425      & 16.926      & 0.766  & 0.200   & 0.656  & 0.000  & 0.964     & 0.000     &  0.757                          \\ \rowcolor{gray!15} \hline
\textbf{ResNet-50 (Ours)~\cite{7}}                                                                     & 78.571                                                        & 98.797      & 95.183      & 98.855 & 94.819  & 98.923 & 94.406 & 97.579    & 97.376    &    95.108                          \\ \rowcolor{gray!15}
\textbf{WideResNet-50 (Ours)~\cite{27}}                                                                 & 78.082                                                        & 99.122      & 96.057      & 98.752 & 94.463  & 98.876 & 94.505 & 98.240    & 97.687    & 95.658                              \\ \rowcolor{gray!15}
\textbf{DenseNet-121 (Ours)~\cite{8}}                                                                  & 74.711                                                        & 98.900      & 94.977      & 98.858 & 94.314  & 98.829 & 94.081 & 97.738    & 97.156    & 95.486                              \\ \rowcolor{gray!15}
\textbf{ShuffleNetV2-x2.0 (Ours)~\cite{42}}                                                             & 71.910                                                        & 98.571      & 92.151      & 98.525 & 92.114  & 98.625 & 92.206 & 97.320    & 97.054    &  98.566                             \\ \rowcolor{gray!15}
\textbf{EfficientNet-B4 (Ours)~\cite{28}}                                                               & 71.759                                                        & 98.328      & 94.677      & 98.831 & 89.953  & 98.993 & 90.865 & 98.321    & 97.968    &   93.068                            \\ \rowcolor{gray!15}
\textbf{MobileNetV3-x1.4 (Ours)~\cite{43}}                                                              & 73.389                                                        & 98.225      & 93.358      & 99.094 & 92.432  & 99.025 & 91.962 & 98.055    & 97.303    &   92.761                            \\ \rowcolor{gray!15}
\textbf{DeiT-S (Ours)~\cite{44}}                                                                        & 79.931                                                       & 97.907      & 91.469      & 99.046 & 93.875  & 99.123 & 93.950 & 98.286    & 97.984    &   91.512                            \\ \rowcolor{gray!15}
\textbf{ViT-B/16 (Ours)~\cite{29}}                                                                      & 79.466                                                       & 98.275      & 92.238      & 99.290 & 94.937  & 99.287 & 95.071 & 98.602    & 98.239    &  93.314                            \\ \rowcolor{gray!15}
\textbf{Swin-Transformer-S (Ours)~\cite{30}}                                                            & 82.785                                                        & 98.271      & 94.377      & 98.156 & 90.708  & 98.350 & 91.044 & 98.178    & 96.766    &  92.306                             \\  \hline
\end{tabular}
\end{center}
\end{spacing}
\vskip -0.13in
\end{table}

\noindent
\textbf{Adversarial Defense on CIFAR-10~\cite{46}.} In the previous section, we compare the model robustness in the presence of adversarial attacks between the vanilla classifier and its GCM counterpart. In this part, we further discuss the performance of the proposed GCM and some existing defending methods. Formally, we report the results of several canonical defending schemes with and without GCM including Unlabeled~\cite{unlabeled}, Limits Uncovering~\cite{uncovering}, Weight Perturbation~\cite{weight}, Fixing Data Aug~\cite{fixing}, Extra Helper~\cite{helper}, Stable Neural ODE~\cite{stable}, Exploring EMA~\cite{ema}, Proxy Distributions~\cite{proxy} and Data Generation~\cite{generation}. We use the pre-trained models from RobustBench~\cite{robustbench} which can be downloaded at this web page \footnote{https://github.com/RobustBench/robustbench}. Only the classification accuracy (ACC) is adopted as the evaluation metric since no attack robustness (AR) values are given in \cite{robustbench}. Besides, all these pre-trained models are designed for defending against PGD attack under the constraint of $L_{\infty}$ norm, and therefore, we only compare these classifiers in the single PGD attack manner. $w$ and $\varepsilon$ are set to 1e20 and 1e-8 respectively.

\setlength{\tabcolsep}{6pt}
\begin{table}[htpb]
\begin{spacing}{1.18}
\begin{center}
\caption{\textbf{Adversarial defense results on CIFAR-10~\cite{46}.} Compared to the current defending methods shown in the following table, GCM is much more stable in handling perturbations with different magnitudes, achieving a superior performance when stated with perturbed images. Values below are classification accuracy on CIFAR-10.}
\label{tab:cifar10}
\vskip -0.02in
\begin{tabular}{c|c|ccccc}
\hline
\multirow{2}{*}{\begin{tabular}[c]{@{}c@{}} CIFAR-10~\cite{46}\\ Testing Set\end{tabular}}  &  \multirow{2}{*}{\begin{tabular}[c]{@{}c@{}}ACC on\\ Clean Images\end{tabular}} & \multicolumn{5}{c}{PGD~\cite{20} $L_{\infty}$}                \\
                                                                                &                                                                                & $\frac{2}{255}$ & $\frac{4}{255}$ & $\frac{6}{255}$ & $\frac{8}{255}$ & $\frac{10}{255}$ \\ \hline

Unlabeled (2019)~\cite{unlabeled}                                                                & 89.69                                                                          & 85.90 & 79.10 & 72.95 & 64.61 & 55.17  \\
Uncovering Limits (2020)~\cite{uncovering}                                                        & 91.10                                                                          & 88.60 & 85.50 & 81.70 & 76.60 & 71.80  \\
Weight Perturbation (2020)~\cite{weight}                                                      & 88.25                                                                          & 86.00 & 82.40 & 77.00 & 72.10 & 67.20  \\
Fixing Data Aug (2021)~\cite{fixing}                                                          & 92.23                                                                          & 88.80 & 84.70 & 78.10 & 70.20 & 63.30  \\
Extra Helper (2021)~\cite{helper}                                                             & 91.47                                                                          & 87.30 & 82.10 & 75.00 & 66.70 & 57.30  \\
Stable Neural ODE (2021)~\cite{stable}                                                        & 93.73                                                                          & 94.00 & 93.80 & 93.30 & 92.40 & 91.90  \\
Exploring EMA (2021)~\cite{ema}                                                            & 91.23                                                                          & 87.80 & 84.30 & 80.00 & 75.80 & 70.00  \\
Proxy Distributions (2021)~\cite{proxy}                                                      & 86.68                                                                          & 84.20 & 81.20 & 77.90 & 73.10 & 68.60  \\
Data Generation (2021)~\cite{generation}                                                          & 87.50                                                                          & 83.70 & 79.30 & 74.90 & 70.60 & 65.60  \\ \rowcolor{gray!15} \hline
\textbf{Unlabeled (Ours)~\cite{unlabeled}}                                                                & 90.30                                                                          & 90.31 & 90.52 & 90.27 & 90.21 & 90.19  \\ \rowcolor{gray!15}
\textbf{Uncovering Limits (Ours)~\cite{uncovering}}                                                        & 90.82                                                                          & 90.50 & 90.40 & 90.70 & 90.80 & 90.30  \\ \rowcolor{gray!15}
\textbf{Weight Perturbation (Ours)~\cite{weight}}                                                      & 88.36                                                                          & 88.50 & 88.60 & 88.10 & 88.00 & 87.90  \\ \rowcolor{gray!15}
\textbf{Fixing Data Aug (Ours)~\cite{fixing}}                                                          & 92.82                                                                          & 92.90 & 92.70 & 92.50 & 92.60 & 92.50  \\ \rowcolor{gray!15}
\textbf{Extra Helper (Ours)~\cite{helper}}                                                             & 91.51                                                                          & 91.50 & 91.60 & 91.79 & 91.73 & 91.52  \\ \rowcolor{gray!15}
\textbf{Stable Neural ODE (Ours)~\cite{stable}}                                                        & 94.09                                                                          & 94.10 & 94.10 & 94.30 & 93.90 & 93.90  \\ \rowcolor{gray!15}
\textbf{Exploring EMA (Ours)~\cite{ema}}                                                            & 91.45                                                                          & 91.30 & 91.30 & 90.90 & 91.00 & 90.70  \\ \rowcolor{gray!15}
\textbf{Proxy Distributions (Ours)~\cite{proxy}}                                                      & 86.91                                                                          & 86.80 & 86.90 & 87.00 & 86.90 & 86.60  \\ \rowcolor{gray!15}
\textbf{Data Generation (Ours)~\cite{generation}}                                                          & 87.91                                                                          & 87.90 & 87.90 & 87.70 & 87.60 & 87.40  \\\hline
\end{tabular}
\end{center}
\end{spacing}
\end{table}

From the results shown in Table~\ref{tab:cifar10}, it can be noticed that the proposed GCM is stable in presence of perturbations with different magnitudes. In particular, when limited the $L_{\infty}$ norm is no more than $\frac{10}{255}$, GCM achieves a superiority with up to 35.02\% improvement in classification accuracy compared to the Unlabeled~\cite{unlabeled} scheme. Generally, FGSM~\cite{1} is less aggressive compared to PGD~\cite{20} attack, and therefore, we eliminate these redundancy experiments on CIFAR-10~\cite{46}.

\subsection{Ablation Study}
\noindent
\textbf{Ablation on magnitude $\varepsilon$.} First, we fix the value of frequency $w$ to 1e20, and the $\varepsilon$ is situated between 1e-8 and 1e-3. It is worth noting that the value of $\varepsilon$ should be sufficiently small in case of introducing intolerable perturbations. We use ResNet-50~\cite{7} and ViT~\cite{29} to test the sensitivity of the model to parameter $\varepsilon$. We report the AR metrics for evaluating the model's robustness.

\setlength{\tabcolsep}{3.3pt}
\begin{table}[htpb]
\begin{spacing}{1.35}
\begin{center}
\caption{\textbf{Ablation on magnitude $\varepsilon$.} The frequency $w$ is set to 1e20. The magnitude of perturbations is limited to $\frac{8}{255}$ with the $L_{\infty}$ norm. Values below are AR results on ImageNet.}
\label{tab:ablation_eps}
\vskip -0.15in
\begin{tabular}{cccccc>{\columncolor{gray!15}}c|ccccc>{\columncolor{gray!15}}c}
\hline
\multicolumn{7}{c|}{ResNet-50~\cite{7}}                             & \multicolumn{6}{c}{ViT-B/16~\cite{29}}                        \\
$\varepsilon$  & 1e-3   & 1e-4   & 1e-5   & 1e-6   & 1e-7   & 1e-8   & 1e-3   & 1e-4   & 1e-5   & 1e-6   & 1e-7   & 1e-8   \\ \hline
FGSM~\cite{1} & 90.29 & 90.92 & 91.02 & 91.02 & 91.01 & \textbf{91.04} & 81.74 & 83.33 & 83.52 & 83.53 & 83.52 & \textbf{83.54} \\
PGD~\cite{20}  & 78.96 & 94.90 & 97.03 & 97.24 & 97.19 & \textbf{97.25} & 77.21 & 95.41 & 96.76 & 96.98 & 96.96 & \textbf{97.00} \\ \hline
\end{tabular}
\end{center}
\end{spacing}
\vskip -0.16in
\end{table}

It can be concluded based on the results shown in Table~\ref{tab:ablation_eps} that GCM is insensitive to the magnitude $\varepsilon$,  leading to better stability and reliability in complicated scenarios. Generally, randomly chosen $\varepsilon$ from 1e-8 to 1e-4 is enough to yield a promising result.

\noindent
\textbf{Ablation on frequency $w$.} We set the value of magnitude $\varepsilon$ to 1e-8, and the frequency $w$ is situated from 1e10 to 1e20. The value of $w$ should be sufficiently great for concealing the gradient direction of the vanilla classifiers. We use the identical models and evaluation metrics as Table~\ref{tab:ablation_eps}.

\setlength{\tabcolsep}{3.3pt}
\begin{table}[htpb]
\begin{spacing}{1.35}
\begin{center}
\caption{\textbf{Ablation on frequency $w$.} The magnitude of perturbations is limited to $\frac{8}{255}$ with the $L_{\infty}$ norm. Values below are AR results on ImageNet.}
\label{tab:ablation_w}
\vskip -0.15in
\begin{tabular}{ccccccc|cccccc}
\hline
\multicolumn{7}{c|}{ResNet-50~\cite{7}}                             & \multicolumn{6}{c}{ViT-B/16~\cite{29}}                        \\
$w$    & 1e10   & 1e12   & 1e14   & 1e16   & 1e18   & 1e20   & 1e10   & 1e12   & 1e14   & 1e16   & 1e18   & 1e20   \\ \hline
FGSM~\cite{1} & 95.07 & 94.52 & 94.91 & \textbf{95.65} & 93.12 & 94.02 & 76.49 & 74.80 & 73.44 & \textbf{76.76} & 74.54 & 74.52 \\
PGD~\cite{20}  & 96.64 & 95.79 & 95.08 & 95.47 & 96.58 & \textbf{97.24} & 76.97 & 74.22 & 75.67 & 75.73 & 75.09 & \textbf{77.92} \\ \hline
\end{tabular}
\end{center}
\end{spacing}
\vskip -0.2in
\end{table}

In Table~\ref{tab:ablation_w}, we notice that the changing tendency of AR metrics along with the frequency $w$ is not such clear when the magnitude of $w$ is sufficiently large. Randomly chosen frequency $w$ from 1e10 to 1e20 yields considerable performance on defending unknown attacks. If not specified, we set $\varepsilon$ and $w$ to 1e-8 and 1e20.

\noindent
\textbf{Ablation on the position of GCM.} GCM can be placed at an arbitrary layer of the deep neural networks. We consider two fashions to place the GCM: (1) Front: set the GCM as the top layer of DNNs; (2) Middle: place GCM immediately after some specific convolutional layers.

\setlength{\tabcolsep}{6pt}
\begin{table}[htpb]
\begin{spacing}{1.35}
\begin{center}
\vskip -0.05in
\caption{\textbf{Ablation on the position of GCM.} The magnitude of perturbations are limited within $\frac{8}{255}$ under the $L_{\infty}$ norm. Block indicates the ConvBlock in ResNet-50~\cite{7}. Values below are ACC on ImageNet. Placing GCM in all layers of the classifier yields the best result.}
\label{tab:ablation_position}
\begin{tabular}{ccccccc>{\columncolor{gray!15}}c}
\hline
\begin{tabular}[c]{@{}c@{}}Location\end{tabular} & \multicolumn{1}{c}{Front} & \multicolumn{1}{l}{Block1} & \multicolumn{1}{c}{Block2} & \multicolumn{1}{l}{Block3} & \multicolumn{1}{c}{Block4} & \multicolumn{1}{l}{Block5} & \multicolumn{1}{c}{All Layers} \\ \hline
\multicolumn{1}{c|}{FGSM~\cite{1}}                              & 77.721                    & 73.892                     & 72.682                     & 70.333                     & 67.284                     & 65.478                     & \textbf{78.572}                         \\
\multicolumn{1}{c|}{PGD~\cite{20}}                               & 77.857                    & 77.051                     & 76.093                     & 73.690                     & 67.542                     & 65.303                     & \textbf{78.574}                         \\ \hline
\end{tabular}
\end{center}
\end{spacing}
\vskip -0.18in
\end{table}

\subsection{Auxiliary Visualization Results}
GCM defends deep neural networks against the adversarial attacks through concealing the gradient's direction of the target models. With the goal of visualizing the modification GCM resulted, we use black points (pixel 0) to indicate the negative direction (\emph{i.e.},-1) of the gradient, and let white points (pixel 1) denote the positive direction (\emph{i.e.},1) of that. Occasionally, the gradient of some pixels may be zero, and we set these positions with a value $\frac{1}{2}$. The perturbations got through FGSM are illustrated as a gray map within each RGB channel, and the result merging these three channels is still shown as a color image. Following figure shows the visualization results.

\begin{figure}[htpb]
\vskip -0.1in
\begin{center}
\centerline{\includegraphics[width=\columnwidth]{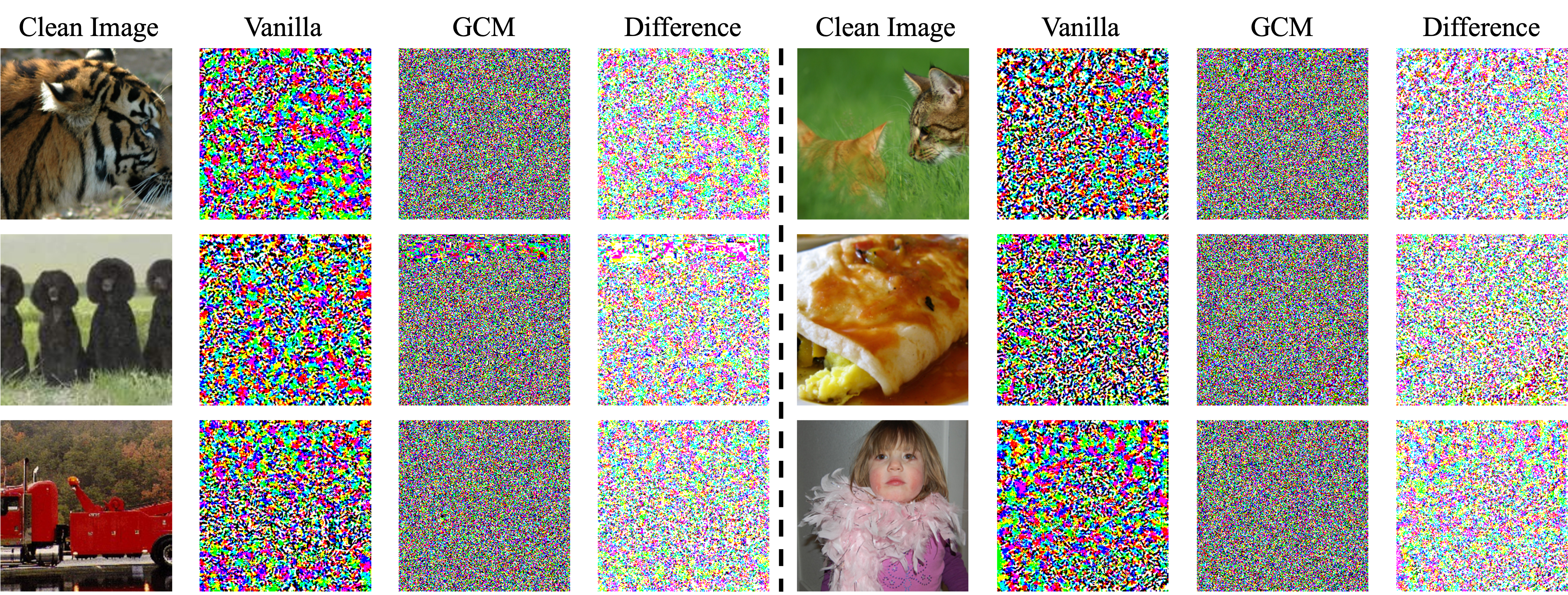}}
\vskip -0.05in
\caption{\textbf{Visualization of the sign of gradients.} It can be seen that the gradient generated by GCM is much more dispersed compared to the vanilla classifier, weakening  the adversarial effects.}
\label{fig:gradient}
\end{center}
\vskip -0.3in
\end{figure}

\section{Conclusion}
\label{conclusion}
In this paper, with the goal of mitigating adversarial effects of DNNs, we propose a novel module termed as GCM to conceal the sign of models' gradient, protecting conventional neural networks from adversarial perturbations. Sufficient experiments on different architectures with and without defending schemes indicate the superior defense ability and stability of our proposed GCM. We are looking forward to see more innovative studies removing obstacles for reliable deployment of DNNs.

\clearpage

{
\small

\bibliographystyle{plain}
\bibliography{ref}

}

\end{document}